\documentclass[letterpaper]{article} 
\usepackage{aaai25}  
\usepackage{times}  
\usepackage{helvet}  
\usepackage{courier}  
\usepackage[hyphens]{url}  
\usepackage{graphicx} 
\usepackage{natbib}  
\usepackage{caption} 
\usepackage{algorithm}
\usepackage{algorithmic}
\usepackage{amsfonts}
\usepackage{MnSymbol}
\usepackage{tabularx}
\usepackage{booktabs} 
\usepackage{array}
\usepackage{multirow}
\newcolumntype{R}[1]{>{\raggedright\arraybackslash}p{#1}}
\usepackage{makecell}
\usepackage{longtable}
\usepackage{xcolor}

\usepackage{newfloat}
\usepackage{listings}
\DeclareCaptionStyle{ruled}{labelfont=normalfont,labelsep=colon,strut=off} 
\floatstyle{ruled}
\newfloat{listing}{tb}{lst}{}
\floatname{listing}{Listing}
\title{Eliciting Causal Abilities in Large Language Models for Reasoning Tasks}
\author{
    Yajing Wang\textsuperscript{\rm 1,2}, Zongwei Luo\textsuperscript{\rm 3,4}\thanks{Corresponding author.}, Jingzhe Wang\textsuperscript{\rm 1}, Zhanke Zhou\textsuperscript{\rm 2}, Yongqiang Chen\textsuperscript{\rm 5}, Bo Han\textsuperscript{\rm 2}
}
\affiliations{
    \textsuperscript{\rm 1} Department of Computer Science, BNU-HKBU United International College\\
    \textsuperscript{\rm 2} TMLR Group, Hong Kong Baptist University\\
    \textsuperscript{\rm 3}  Artificial Intelligence and Future Networks IAIFN, Beijing Normal University at Zhuhai\\ 
    \textsuperscript{\rm 4} Guangdong Provincial Key Laboratory of IRADS\\
    \textsuperscript{\rm 5} The Chinese University of Hong Kong \\
    \{wangyajing, wangjingzhe\}@uic.edu.cn, lzwqhk@outlook.com, \\
    \{cszkzhou,bhanml\}@comp.hkbu.edu.hk, yqchen@cse.cuhk.edu.hk
}

\usepackage{bibentry}

\begin{document}

\maketitle

\begin{abstract}
Prompt optimization automatically refines prompting expressions, unlocking the full potential of LLMs in downstream tasks. However, current prompt optimization methods are costly to train and lack sufficient interpretability. This paper proposes enhancing LLMs' reasoning performance by eliciting their causal inference ability from prompting instructions to correct answers. Specifically, we introduce the \textit{Self-Causal Instruction Enhancement} (SCIE) method, which enables LLMs to generate high-quality, low-quantity observational data, then estimates the causal effect based on these data, and ultimately generates instructions with the optimized causal effect. In SCIE, the instructions are treated as the treatment, and textual features are used to process natural language, establishing causal relationships through treatments between instructions and downstream tasks. Additionally, we propose applying \textit{Object-Relational} (OR) principles, where the uncovered causal relationships are treated as the inheritable class across task objects, ensuring low-cost reusability. Extensive experiments demonstrate that our method effectively generates instructions that enhance reasoning performance with reduced training cost of prompts, leveraging interpretable textual features to provide actionable insights.

\end{abstract}

%
\begin{links}
     \link{Code}{https://github.com/dsubuntu/SCIE}
\end{links}

\section{Introduction}
One major remaining challenge for Large Language Models (LLMs) is their insufficient reasoning capabilities \cite{dziri2024faith, cao2024envisioning}. Current LLMs perform well on System-1 tasks but face limitations in handling System-2 problems \cite{bengio2019system}. Prompting-based methods \cite{lester2021power, liu2023pre} aim to enable LLMs to understand input prompts and adapt to the downstream tasks through the design and crafting of prompts, becoming a focal point of interest among researchers in recent years. Compared to fine-tuning methods \cite{howard2018universal, dong2019unified, lewis2019bart}, prompting methods do not require substantial computational resources and time to retrain the model, allowing for the development of more generalized solutions using the original pre-trained model \cite{li2023deepinception}.

\begin{figure}[t]
\centering
\includegraphics[width=\columnwidth]{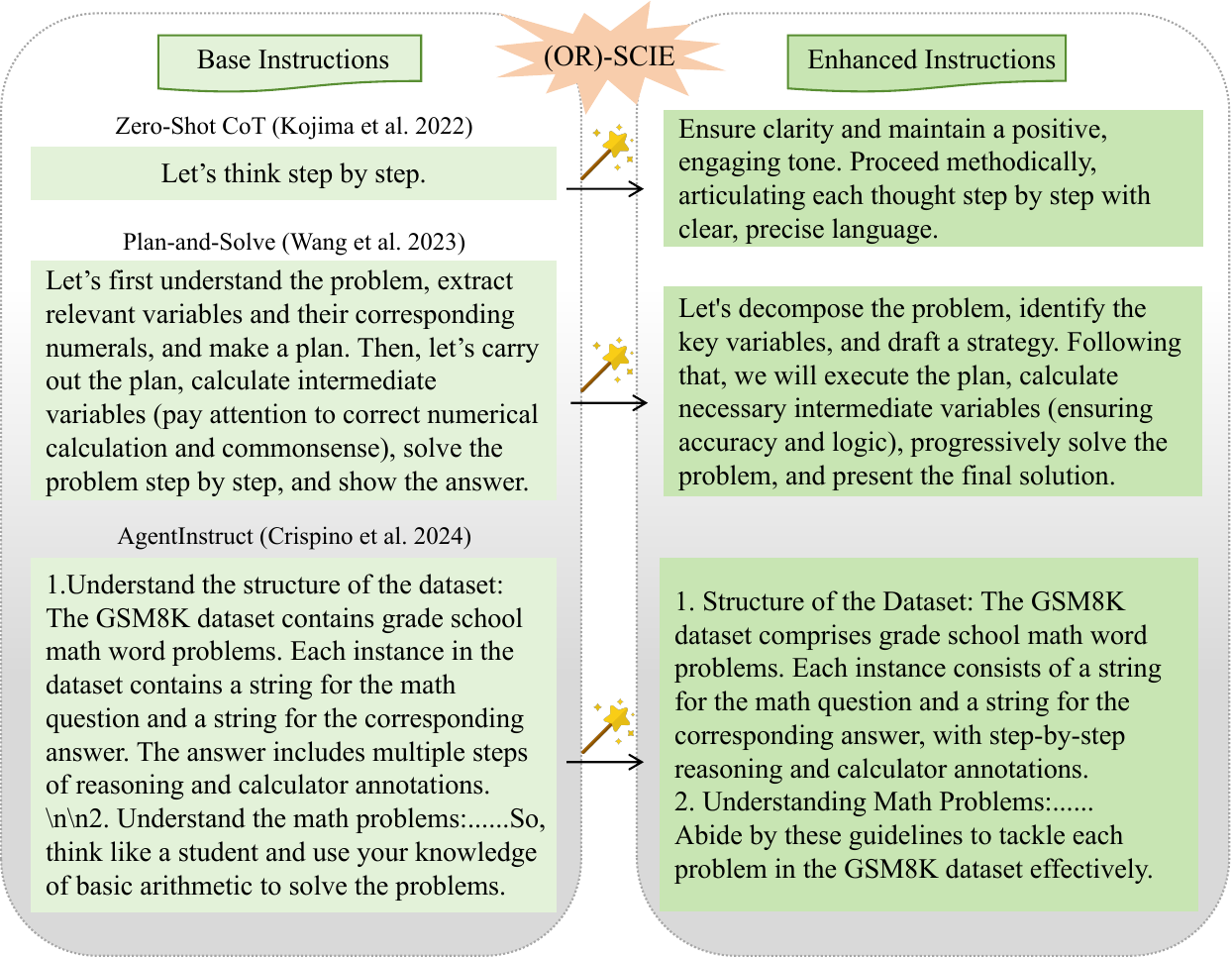}
\caption{Illustrative examples demonstrating the purpose of our proposed \textit{(Object-Relational) Self-Causal Instruction Enhancement} method.}
\vspace{-20pt} 
\label{Fig1}
\end{figure}

Many exciting prompting methods have emerged, such as Chain of Thought (CoT) \cite{wei2022chain}, Zero-Shot CoT \cite{kojima2022large}, among others. The prompting instructions in these methods are typically designed by humans, even introducing some noise \cite{zhou2024can}. This leads us to question: are these language expressions of the prompts the best to trigger LLMs? Prompt optimization methods \cite{chang2024efficient} refine and enhance prompts for LLMs, to improve LLMs' performance on downstream tasks. However, current prompt optimization methods face challenges related to training costs and interpretability. Gradient-based approaches, such as APE \cite{zhou2022large}, APO \cite{pryzant2023automatic}, and OPRO \cite{Yang2023LargeLM}, require substantial training costs to obtain gradient information. While gradient-free methods like GPS \cite{xu2022gps} and GrIPS \cite{prasad2023grips}, which rely on editing and searching, also lack the interpretability to be understood from a human intuitive perspective during the process.

Causal abilities represent higher-level cognition that transitions from System 1 to System 2 \cite{bengio2019system}, and LLMs demonstrate potential in causal reasoning tasks \cite{kiciman2023causal}. We propose enhancing the reasoning ability of LLMs by eliciting their ability for causal inference and aim to design a method to reduce training costs while improving interpretability. Furthermore, inspired by the core idea of meta-prompting abstracting a certain structure or pattern of prompts that exhibits good generalizability, we propose regarding the uncovered causal relationships as an abstract meta-template to guide the generation of prompts.

This paper proposes the \textit{Self-Causal Instruction Enhancement} (SCIE) method. Given a basic prompting instruction and several correct annotations for the corresponding downstream task results, SCIE allows LLMs to perform causal estimation and optimization on the given instruction, resulting in better reasoning performance. Moreover, inspired by meta prompting, we employ \textit{Object-Relational} (OR) thinking, enabling new downstream tasks to inherit the uncovered causal relationships. This approach facilitates easier and more cost-effective optimization of the instructions. As illustrated in Figure~\ref{Fig1}, given any input instruction and prompting method, the (OR)-SCIE method generates enhanced instructions on the reasoning performance, such as the accuracy of LLMs. 

We summarize the contributions of this paper as follows:
\begin{itemize}
    \item To the best of our knowledge, this is the first work that enhances the prompts of LLMs for reasoning tasks from a causal perspective. The proposed SCIE method elicits the causal abilities of LLMs to improve their reasoning ability and provide interpretability.
    
    \item Inspired by the theory of causal identification, we generate high-quality, low-quantity observational data, addressing the need for observational data for causal inference on LLM prompts and downstream task outcomes.
    
    \item The uncovered causal relationships between instructions and task outcomes can be regarded as a class and reused in other downstream tasks that satisfy the OR relationship. The experiment shows new tasks that inherit the corresponding causal relationships through the OR model demonstrate improved performance.
\end{itemize}

\section{Preliminaries}
\subsection{Causal Estimand}

To quantify the causal effect, we need to identify the causal estimands. The causal effect for an individual, referred to as the Individual Treatment Effect (ITE) \cite{holland1986statistics}, is challenging to identify due to the counterfactual problem. However, we can estimate the overall average level, namely, Average Treatment Effect (ATE) \cite{rubin1974estimating}:
\begin{equation}
    \mathrm{ATE}=\mathbb{E}[Y(1)-Y(0)],
\end{equation}
where the ATE represents the average difference in potential outcome variables $(Y)$ between the treatment group $(1)$ and the control group $(0)$.


\subsection{Identification Assumptions for
Causal Inference}
To perform causal inference using a causal estimand, three identification assumptions need be satisfied: ignorability, positivity, and consistency \cite{feder2022causal}.
\subsubsection{Ignorability.} Ignorability, also known as unconfoundedness, refers to the condition where, for groups with the same values of covariates, the assignment of the treatment is independent of the potential outcome variables:
\begin{equation}
    T \upmodels Y(t) \mid X, \quad \forall t \in\{0,1\},
\end{equation}
where $X$ is observed variables (including confound variables),  $T$ and $Y$ are the treatment and the potential outcome separately, and $t$ means the value of $T$. In other words, this assumption requires that we observe all confounding variables and that there is sufficient variation in $X$.
\subsubsection{Positivity.} Positivity refers to the condition that, for any given observed variable $X$, the assignment of the intervention $T$ has a probability between 0 and 1:
\begin{equation}
0<\operatorname{Pr}(T=1 \mid X=x)<1, \quad \forall x ,
\end{equation}
requiring the assignment of treatments be random, meaning that each unit has a non-zero probability of being treated.
\subsubsection{Consistency.} The Consistency Assumption, also known as the Stable Unit Treatment Value Assumption (SUTVA), states that the potential outcome $Y$ of any unit is not influenced by the treatment $T$ applied to other units. Additionally, for each unit, there are no different forms or versions of any given $T$ that could lead to different $Y$:
\begin{equation}
T=t \Leftrightarrow Y(t)=Y, \quad \forall t \in\{0,1\}.
\end{equation}

This assumption requires that each treatment be clearly defined and that the potential outcomes resulting from the treatment are stable.

\begin{figure*}[h]
\centering
\includegraphics[width=\textwidth]{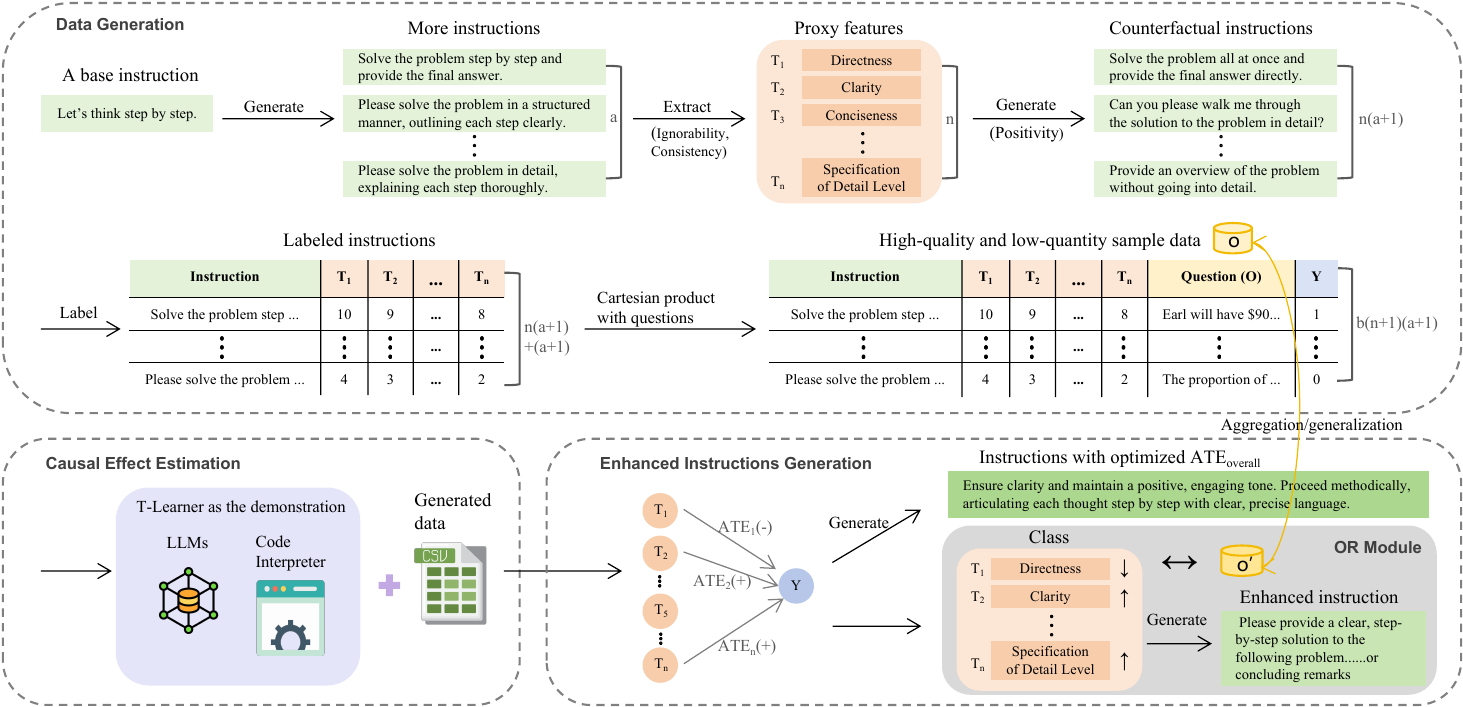}
\caption{The overall process of \textit{(Object-Relational) Self-Causal Instruction Enhancement} includes Data Generation, Causal Effect Estimation, Enhanced Instructions Generation, and optional OR Module.}
\vspace{-19pt} 
\label{Fig2}
\end{figure*}

\section{Self-Causal Instruction Enhancement}

This paper aims to enhance the reasoning performance of LLMs in downstream reasoning tasks by estimating and optimizing the causal effects of prompting instructions. A natural thought for causal inference of prompts and task outcomes in LLMs is the potential outcome framework \cite{rubin1974estimating} since we can guide LLMs in generating
data that align as closely as possible with requirements. In our causal effect estimation method, the instructions serve as the treatment $T$, and the correctness of the results in downstream tasks serves as the potential outcome $Y$. This allows us to further enhance the instructions based on the estimated causal relations. The causal effect is: 
\begin{equation}
\begin{aligned}
    \mathrm{ATE} &= \mathbb{E}[Y(1) - Y(0)] \\
    &= \mathbb{E}[Y(1)] - \mathbb{E}[Y(0)] \\
    &= \mathbb{E}[Y \mid T=1 ] - \mathbb{E}[Y \mid T=0 ] .
\label{equ}
\end{aligned}
\end{equation}

For simplicity without loss of generality, we assume $T$ is expressed in binary form here. The values of $Y$ are 1 and 0, where 1 indicates a correct outcome, and 0 indicates an incorrect outcome. In other words, we represent the reasoning ability of LLMs through the causal effect of instructions on the correctness of reasoning task outcomes. To improve the reasoning capabilities of LLMs in downstream tasks, we simply need to identify the instructions that maximize the causal effects of the prompts to the correctness of the task.

Theoretically, the counterfactual $Y(t)$ can be defined for any treatment $t$, but it might be restricted to the representation of the text \cite{feder2022causal}. Methods for handling high-dimensional text as treatment can be divided into two categories. The first category focuses on producing interpretable features of the text \cite{pryzant2018deconfounded, kunzel2019metalearners, maiya2021causalnlp}. The second category involves extracting latent properties of the text during causal effect estimation \cite{wood2018challenges, pryzant2021causal}, but they typically require a proposed causal model and rely on the validity of the model. Escaping from relying on expert knowledge and for the sake of interpretability, our SCIE will extract features of interest (referred to as proxy features, whose values are typically inferred from text with classifiers, lexicons, or topic models \cite{pryzant2018deconfounded, kunzel2019metalearners, maiya2021causalnlp}) from high-dimensional prompting instructions, preparing for the causal effect estimation as treatments.

In general, the potential outcomes framework is based on observational data under three key assumptions. Although treating LLMs does not raise ethical concerns, conducting a large number of randomized controlled experiments on LLMs is inconvenient and costly to replicate. Therefore, we propose generating high-quality (satisfying the three assumptions) and low-quantity observational data and then estimating causal effects based on the data. Then, based on the uncovered causal effects, we ask the LLM to generate enhanced instructions that have a stronger causal effect on the correctness of the results in downstream tasks. Additionally, the uncovered causal model can be regarded as a meta-prompting pattern, which can be inherited according to the OR approach, thereby improving the method's cost-effective reusability. The overall process of our method is shown in Figure~\ref{Fig2}. We will illustrate in details for the (OR)-SCIE method in the following sections.

\subsection{High-quality Observational Data Generation}
In this section, we will explain how to generate observational data that satisfies the three identification assumptions in causal inference while ensuring that the resulting structured data is manageable for causal effect estimation.

As shown in the Data Generation part of Figure~\ref{Fig2}, based on an instruction for LLMs to complete a task, which can be either manually constructed or automatically generated based on previous research, we generate $a$ different instructions from the given instruction using LLMs, such as forward mode and reverse mode in APE \cite{zhou2022large}. Assuming we are interested in the textual features of the instructions, as these features are relatively generic for the input instruction and facilitate subsequent processing, we have LLMs exhaustively enumerate all $n$ textual features $\{T_1, T_2, \ldots, T_n\}$ that influence the results of downstream tasks for these $(a+1)$ instructions, and these proxy features must be independent of each other (ignorability). The proxy features must be described in detail and be consistent without version bias (consistency). Next, we have the LLMs generate counterfactual instructions based on each proxy feature, resulting in a total of $n(a+1)$ counterfactual instances, ensuring that each proxy feature has both the probability of being treated and not being treated (positivity). 

To label these proxy features as numerical or categorical data, we use the method by leveraging the scoring capabilities of LLMs for annotation \cite{liu2024discovery}. Subsequently, we randomly select $b$ data from the training set of the LLMs' downstream reasoning tasks. These $b$ questions are combined with the $(n+1)(a+1)$ instructions (including the counterfactual and original instructions) using the Cartesian product, resulting in $b(n+1)(a+1)$ instruction-question instances. These instances are then input into the LLMs, and the potential outcome $Y$ indicating the correctness of the answers can be obtained to finish the data generation process.

\begin{figure}[h]
\centering
\includegraphics[width=\columnwidth]{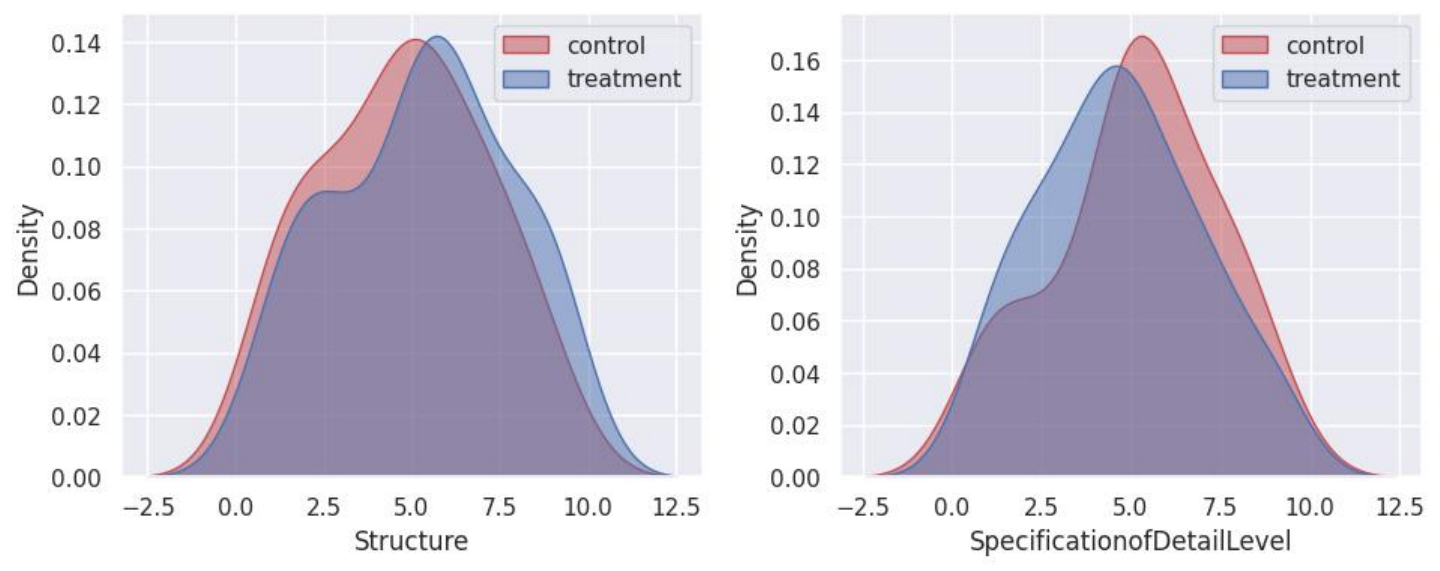}
\caption{Probability density distributions of the proxy features ``Structure" and ``Specification of Detail Level" with ``Directness" as the proxy treatment.}
\vspace{-7pt} 
\label{Fig3}
\end{figure}

We use an example to illustrate the desired observational data for causal effect estimation. In this example, the LLM is GPT-4o mini, and the task is GSM8K, with the given base instruction being Zero-Shot CoT \cite{kojima2022large}. According to the data generation method, we derive $8$ proxy features: ``Directness", ``Clarity", ``Conciseness", ``Actionability", ``Tone", ``Structure", ``Specification of Detail Level", and ``Emphasis on Process". We specify generating $9$ different instructions from the base instruction, resulting in $(8+1)(9+1)=90$ instructions. Each of these proxy features is sequentially treated as the treatment variable with the treated and controlled value (1 and 0), while the remaining features serve as covariates. We binarize the current treatment ``Directness" and plot the probability density distributions of the other proxy features over the current treatment. We show two examples of the distributions for ``Structure" and ``Specification of Detail Level" in Figure~\ref{Fig3}.

Ideally, the overlap of the distributions between the controlled and treated groups of the treatment variable can be considered as perfect observational data for causal effect estimation. As shown in Figure~\ref{Fig3}, although there are differences, the distributions of the controlled and treated groups share significant commonalities, making them suitable for causal effect estimation. We also attempt to further adjust the data using Propensity Score Matching (PSM) \cite{rosenbaum1983central, dhawan2024end}, but find that this approach results in poorer outcomes. We attribute this to the reduction in data volume post-PSM, which adversely affects the causal effect estimation and hinders the generation of better instructions, so we discard the PSM step.

\subsection{Estimating Causal Effect with LLMs}

In the Data Generation step, we prepare the data for causal effect estimation. In this section, our goal is to perform causal effect estimation on each proxy treatment to uncover the causal relationships between the treatments and the correctness of LLM downstream task results. 

During the causal effect estimation, for each proxy feature $T_i \in \{T_1, T_2, \ldots, T_n\}$ considered as a treatment in turn, its causal effect is estimated:
\begin{equation}
\begin{aligned}
    \mathrm{ATE_i}
    &= \mathbb{E}_x[\mathbb{E}[Y(1) \mid X] - \mathbb{E}[Y(0) \mid X]] \\
    &= \mathbb{E}_x[\mathbb{E}[Y \mid T_i = 1, x] - \mathbb{E}[Y \mid T_i = 0, x]],
\label{equ5}
\end{aligned}
\end{equation}
where $X$ represents covariates (proxy features other than the current proxy treatment), and $x$ represents a specific value of $X$. $E_x$ indicates the expectation over all values of $X$. 

Ordinary LLMs typically struggle with table data processing. When LLMs are asked to calculate ATE based on Equation~\ref{equ5}, common responses include problem-solving steps or a piece of code without answers. To solve this problem, we employ the open interpreter \cite{OpenInterpreter} for ATE estimation. It contains a built-in code interpreter that can generate and execute code based on prompts and return results. The reason for using LLMs to estimate causal effects is that, subsequently, to generate instructions with larger $\mathrm{ATE_{overall}}$, the LLM needs to estimate correct ATE and understand the ATE calculation process. To enable LLMs to accurately estimate the $\mathrm{ATE_i}$, we utilize the in-context learning strategy \cite{brown2020language}, providing the relevant code along with $\lceil i/2 \rceil$ ATE results as the demonstration to the LLMs and ask LLMs estimating the complete $i$ ATE results. This process is shown in the Causal Effect Estimation part of Figure~\ref{Fig2}. The idea of this part is to leverage expert knowledge to teach LLMs how to perform causal effect estimation, enabling them to apply the learned knowledge to excel in their strength of language generation.

The meta-learners, including T-Learner, S-Learner \cite{kunzel2019metalearners, maiya2021causalnlp}, are effective methods in potential outcomes framework to estimate causal effect. The T-Learner is considered as the example for in-context learning, as the treatment effect between our control and treatment groups differs significantly in our data (the counterfactual instruction process), and has low selection bias. The T-Learner employs the base learner (e.g., the supervised learning or regression estimator) to separately estimate the control and treatment group functions:
\begin{equation}
\begin{aligned}
& \mu_0(x)=\mathbb{E}[Y(0) \mid X=x], \\
& \mu_1(x)=\mathbb{E}[Y(1) \mid X=x].
\end{aligned}
\end{equation}

The difference of these two estimates $\hat{\mu}_1(x)$ and $\hat{\mu}_0(x)$ denotes the estimation of the ATE using T-learner:
\begin{equation}
\hat{\tau}(x)=\hat{\mu}_1(x)-\hat{\mu}_0(x).
\end{equation}

The S-Learners are considered as the other baseline for evaluating the ATE estimated by LLMs. Detailed experimental results are provided in the following section.
    


\subsection{Enhanced Instructions Generation}
By estimating the ATE of different proxy treatments, we obtain insights into how instructions causally influence the correctness of LLM downstream task outcomes through proxy features. Next, we aim to adjust the instructions based on the uncovered causal relationships to optimize the causal impact of instructions on the correctness of LLM downstream task outcomes. Considering the independence among the proxy features, we define the overall ATE of instructions on the correctness of LLM downstream task outcomes as:
\begin{equation}
\mathrm{ATE_{overall}} = \frac{1}{n} \sum_{i=1}^n \mathrm{ATE_i}.
\end{equation}

We intend to achieve the optimized $\mathrm{ATE_{overall}}$ not by involving gradient-based optimization concepts but by allowing the LLM to generate instructions that optimize $\mathrm{ATE_{overall}}$ based on the uncovered causal relationships, which leverages the LLM’s capabilities to encode and decode natural language, shown in the Enhanced Instructions Generation part of Figure~\ref{Fig2}. 

Assuming that the LLM can generate instructions with varying degrees of proxy features by adjusting the value of the treatment, a larger ITE can be obtained with a higher probability. We will next prove this point. 

The Individual Treatment Effect (ITE) is represented as:
\begin{equation}
\mathrm{ITE_{i}}^{(j)} = Y_{i}^{(j)}(1)-Y_{i}^{(j)}(0),
\end{equation}
where $\mathrm{ITE_{i}}^{(j)} $ and $Y_{i}^{(j)}$ represent treatment effect and outcome of an individual corresponding to $T_i$, respectively.

The $\mathrm{ATE_i}$ can be represented as:
\begin{equation}
\mathrm{ATE_{i}} = \frac{1}{m} \sum_{j=1}^m \mathrm{ITE_{i}}^{(j)}.
\end{equation}
That is, the ATE is the average of all individuals' ITE, where $m$ denotes the number of individuals.

\begin{itemize}
\item When $\mathrm{ATE_i}>0$, which means that the effect of $T_i$ is positive. According to the law of large numbers, the overall average effect is the average of individual effects. Therefore, it can be inferred that for an individual $j$, the likelihood that $Y_i^{(j)}(1)$ is greater than $Y_i^{(j)}(0)$ is higher, which implies that adjusting $T_i$ from $0$ to $1$ will increase the value of $\mathrm{ITE_i}$. 

\item Similarly, when $\mathrm{ATE_i}<0$, it can be proven that adjusting $T_i$ from $1$ to $0$ will increase the value of $\mathrm{ITE_i}$.

\item When $\mathrm{ATE_i}=0$, it means that $T_i$ has no significant effect on the overall positive outcome, so the adjustment of $T_i$ can be ignored in this case.
\end{itemize}

Therefore, based on the discussion of adjusting $T_i$ to increase $\mathrm{ATE_i}$, it is possible for the LLM to generate instructions that improve the $\mathrm{ATE_{overall}}$, under the consistency assumption that different proxy treatments affect potential outcomes independently. It is noteworthy that though we aim to obtain instructions with maximum $\mathrm{ATE_{overall}}$, the LLM will give solutions to generate optimized instructions. However, solutions generated by our method show enhanced performance compared to the base instructions.

\subsection{The OR Module}
The object-relational model is a database model proposed to combine the characteristics of relational databases with object-oriented programming. It extends the traditional relational database model by supporting complex data types, inheritance, and other object-oriented features \cite{carey1997bucky}. In the OR approach, there are two common types of object relationships: aggregation and generalization. Aggregation represents associations between collections of objects, emphasizing the whole-part relationship. Generalization relationships among objects allow for the creation of hierarchies where classes can inherit attributes and behaviors from other classes of objects. Objects that satisfy specific relationships can extend or inherit the attributes and methods of existing classes, offering flexibility and reusability. This enables more efficient management of complex data.

The causal model is regarded as a framework that can explicitly represent causal dependencies and allow for automatic reasoning about these dependencies \cite{jensen2021improving}. Applying the OR model in causal inference leads to a more expressive and flexible causal representation of a complex world \cite{jensen2021improving, lee2021network, wang2024exploring}. Based on the above argument, we model different tasks and uncover causal relationships using the OR model to achieve easy reuse of instruction enhancement. As shown in the OR Module part of Figure~\ref{Fig2}, we extract the uncovered causal relationships from the LLM's explanations, which are identified by the LLM as having a significant impact on potential outcomes when generating instructions with optimized ATE. Usually, LLMs encourage the positive impact of features to guide the LLM's-self in generating better instructions. Then, objects (such as different tasks, $O$ and $O^{\prime}$ in Figure~\ref{Fig2}) that have aggregation or generalization relationships can directly inherit from this class, and they can adjust instructions based on the positive or negative impact of the proxy features. This class can be regarded as the meta-template, which can be reused by other objects. 


\section{Experiments and Results}

\begin{figure*}[h]
\centering
\includegraphics[width=\textwidth]{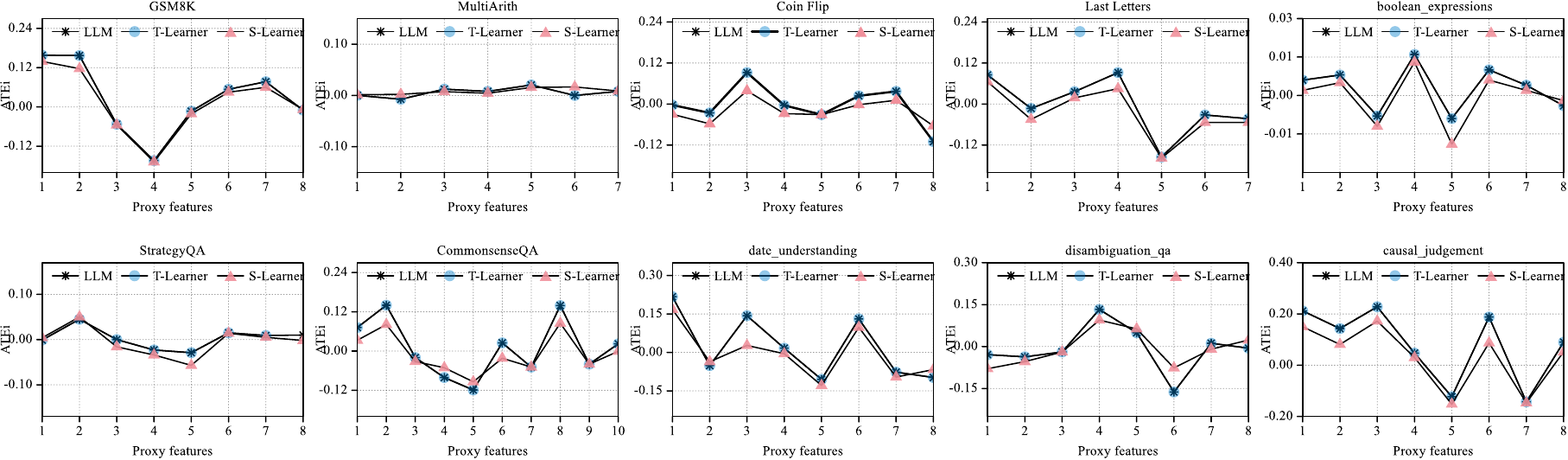}
\caption{The causal effect estimation with the LLM (asterisk points), T-Learner (blue circular points), and S-Learner (red triangle points). The X-axis represents the $n$ proxy features for each reasoning task (indicated by numerical ticks), used as treatments when calculating $\mathrm{ATE_i}$. The Y-axis represents the corresponding ATE values of treatments.}
\vspace{-15pt}
\label{Fig4}
\end{figure*}

While the enhanced metric could be changed to others, such as the certainty of downstream task answers, perplexity, answer length, etc., this paper focuses on the accuracy of reasoning tasks for LLMs. To validate the effectiveness of (OR)-SCIE on the accuracy of LLMs in reasoning tasks, we set up the experiments with representative reasoning tasks and datasets, LLMs, and baselines.

\subsubsection{Reasoning tasks and datasets.} We evaluate ten common datasets across four categories of reasoning tasks for the experiment. (1) Arithmetic reasoning: GSM8K \cite{cobbe2021training} and MultiArith \cite{roy2015solving}. (2) Commonsense reasoning: StrategyQA \cite{geva2021did} and CommonsenseQA \cite{talmor2019commonsenseqa}. (3) Symbolic reasoning: Coin Flip \cite{wei2022chain}, Last Letter Concatenation \cite{wei2022chain} and Boolean Expressions \cite{suzgun2023challenging}. (4) Other logical reasoning: Causal Judgement, Date Understanding, and DisambiguationQA from Big Bench Hard (BBH) \cite{suzgun2023challenging}. For datasets like GSM8K, where the training and test sets are pre-defined, we perform random sampling on the training set and evaluate using the test set. For datasets without predefined training and test sets, we exclude the sampled data used in the SCIE process during testing on the reasoning tasks.

\subsubsection{Models.} The experiments in this paper will evaluate inference tasks on several commonly used LLMs, including GPT-3.5 Turbo \cite{OpenAI2022}, GPT-4o mini \cite{achiam2023gpt} and Llama-3-70b \cite{dubey2024llama}. The (OR)-SCIE process is designed to utilize more powerful LLMs like GPT-4o \cite{achiam2023gpt} whenever possible, aiming to enhance the performance of a student model (processing downstream tasks) by activating the causal ability from a teacher (good at causal inference) model.

\subsubsection{Baselines.} Three base instructions and their corresponding prompting methods are used as baselines: Zero-Shot CoT \cite{kojima2022large}, Plan-and-Solve Prompting \cite{wang2023plan}, and AgentInstruct \cite{crispinoagent}. Among them, Zero-Shot CoT and Plan-and-Solve Prompting use a two-stage prompting method \cite{kojima2022large}. In the first stage, instructions like ``Let’s think step by step." are added after the specific question to generate the required CoT. In the second stage, both the question and the CoT generated in the first stage are input into the LLMs to obtain the final answer. AgentInstruct is a one-stage prompting method where the same instruction is input before each question for the same task to obtain the answer. Additionally, we also note the work using LLMs as optimizers \cite{Yang2023LargeLM}, which perform prompt optimization on Zero-Shot CoT, so we will also specifically compare our method with theirs. 


\subsection{Evaluation of Estimating Causal Effect with LLMs}

Having LLMs estimate ATE within the SCIE framework is necessary because, to enable the LLM to generate instructions with the optimized ATE, the LLM needs to understand the principle of obtaining ATE. While estimating ATE with LLMs is not mandatory for the OR module. As long as there is a class reflecting the real causal relationships, the LLM can generate enhanced instructions based on the class.

Comparative experiments are conducted to verify whether LLMs are competent to estimate ATE through our generated data. We use Zero-Shot CoT as the base instruction and generated high-quality observational data within SCIE, where $a=9$, $b=5$, and $n=8$, which is automatically generated by GPT-4o mini (using this setting in the following experiments if not specified). Based on the generated data, we feed the LLM with the code of T-Learner and $4$ answers of $\mathrm{ATE_i}$, letting it estimate all values of $\mathrm{ATE_i}$. We conduct experiments on each reasoning task and compare the results with the S-Learner to further assess the accuracy of the causal effects estimated by the LLM.

Figure~\ref{Fig4} shows the ATE estimation results on different reasoning tasks. We can observe that the LLM has fully mastered the use of the T-Learner for ATE estimation, as all the results it generated, including the points with unknown values, aligning perfectly with those produced by the T-learner. Although there is a slight difference between the ATE estimation and the S-Learner, the overall trend of the LLM's results is consistent, and the difference is minor. This indicates the ATE estimation by the LLM is robust and also reflects the good quality of our generated observational data. It is worth noting that although some ATE results appear close to zero, this does not imply that the treatment variable has no causal effect on the task outcome correctness. On the contrary, the LLM will provide explanations and synthesize the impact of all proxy treatments to offer suggestions for enhancing the instructions, even slightly.

\subsection{Evaluation of SCIE}

\subsubsection{SCIE on Zero-Shot CoT.}
We apply SCIE on the Zero-Shot CoT across reasoning tasks and obtain corresponding enhanced instructions with the LLM explanations (see Appendix A). Table~\ref{Tab1} shows the accuracy of Zero-Shot CoT and SCIE Zero-Shot CoT on reasoning tasks using GPT-3.5 turbo (results on more LLMs can be found in Appendix B).

\begin{table*}[h]
\centering
\resizebox{\textwidth}{!}{
\renewcommand{\arraystretch}{0.1}
\begin{tabular}{c c c c c c}
\midrule[1pt] 
\multirow{2}{*}{} & \multicolumn{2}{c}{Arithmetic reasoning} & \multicolumn{3}{c}{Symbolic reasoning} \\ \cmidrule(r){2-3} \cmidrule(r){4-6}
 & GSM8K & MultiArith & Coin Flip & Last Letters & Boolean Expressions \\ \midrule 
Zero-Shot CoT & 75.5 & 93.8 & 76.8 & 86.5 & 78.4 \\ \cmidrule(r){2-3} \cmidrule(r){4-6}
SCIE Zero-Shot CoT  & \textbf{77.3} & \textbf{95.2} & \textbf{78.2} & \textbf{87.6} & \textbf{79.6} \\ \midrule[1pt] 
\multirow{2}{*}{} & \multicolumn{2}{c}{Common-sense reasoning} & \multicolumn{3}{c}{Other logical reasoning} \\ \cmidrule(r){2-3} \cmidrule(r){4-6}
& StrategyQA& CommonsenseQA & Causal Judgement& Date Understanding &  DisambiguationQA \\ \midrule 
Zero-Shot CoT & 65.5 & 71.7 & 55.5 & \textbf{71.8} & 61.2 \\ \cmidrule(r){2-3} \cmidrule(r){4-6}
SCIE Zero-Shot CoT  & \textbf{71.1} & \textbf{72.0} &\textbf{59.3} & 71.0 & \textbf{61.6}\\ 
\midrule[1pt] 
\end{tabular}}
\caption{Accuracy on reasoning tasks with and without SCIE. Two-stage prompting is employed, with all other experimental settings remaining identical except for the instructions. Values in bold denote better accuracy.}
\vspace{-5pt}
\label{Tab1}
\end{table*}

\begin{table*}[h]
\centering
\resizebox{\textwidth}{!}{
\renewcommand{\arraystretch}{1}
\begin{tabular}{@{} p{1.5cm} p{5cm} p{9cm} p{1.5cm} @{}}
\toprule
\textbf{\makecell[c]{Method}}&\textbf{\makecell[c]{Interpretability}} & \textbf{Instruction}& \textbf{Accuracy}\\ \midrule
\makecell[c]{OPRO} &  \makecell[c]{$\backslash$} & Analyze the given information, break down the problem ... carefully consider the problem’s context for an efficient solution (referenced from Table 4 in \cite{Yang2023LargeLM}). &  \makecell[c]{77.7}\\ \hline
\makecell[c]{SCIE} & To maximize the overall Average Treatment Effect ... (Please refer to Appendix A for a full explanation.) & Ensure clarity and maintain a positive, engaging tone. Proceed methodically, articulating each thought step by step with clear, precise language. & \makecell[c]{\textbf{78.6}} \\ \bottomrule
\end{tabular}}
\caption{Comparison between OPRO and SCIE, both following the ``Q-end" setting \cite{Yang2023LargeLM} which means the instruction is added after the original question, rather than the two-stage prompting \cite{kojima2022large}.}
\vspace{-10pt}
\label{Tab2}
\end{table*}

As shown in Table~\ref{Tab1}, SCIE effectively enhances instructions for most reasoning tasks. However, Date Understand is an exception. We attribute this to a potential bias in the ATE estimation process, as also evident in Figure~\ref{Fig4}. Furthermore, our method does not require extensive training and provides interpretability. We also observe that, aside from the bias in estimation process, if the ATE values exhibit significant fluctuations, the LLM will adjust the instructions based on more pronounced proxy features, resulting in greater performance improvements. For example, the Causal Judgement task shows more pronounced ATE variations (see Figure~\ref{Fig4}) compared to other tasks, and the LLM emphasizes crafting the instruction to maximize the factors that have the most positive effect in the explanations (see the LLM explanations in Appendix A), and consequently, it achieves a higher accuracy improvement relative to the other tasks. This observation aligns with the design principles of our method.

\begin{figure}[h]
\centering
\includegraphics[width=\columnwidth]{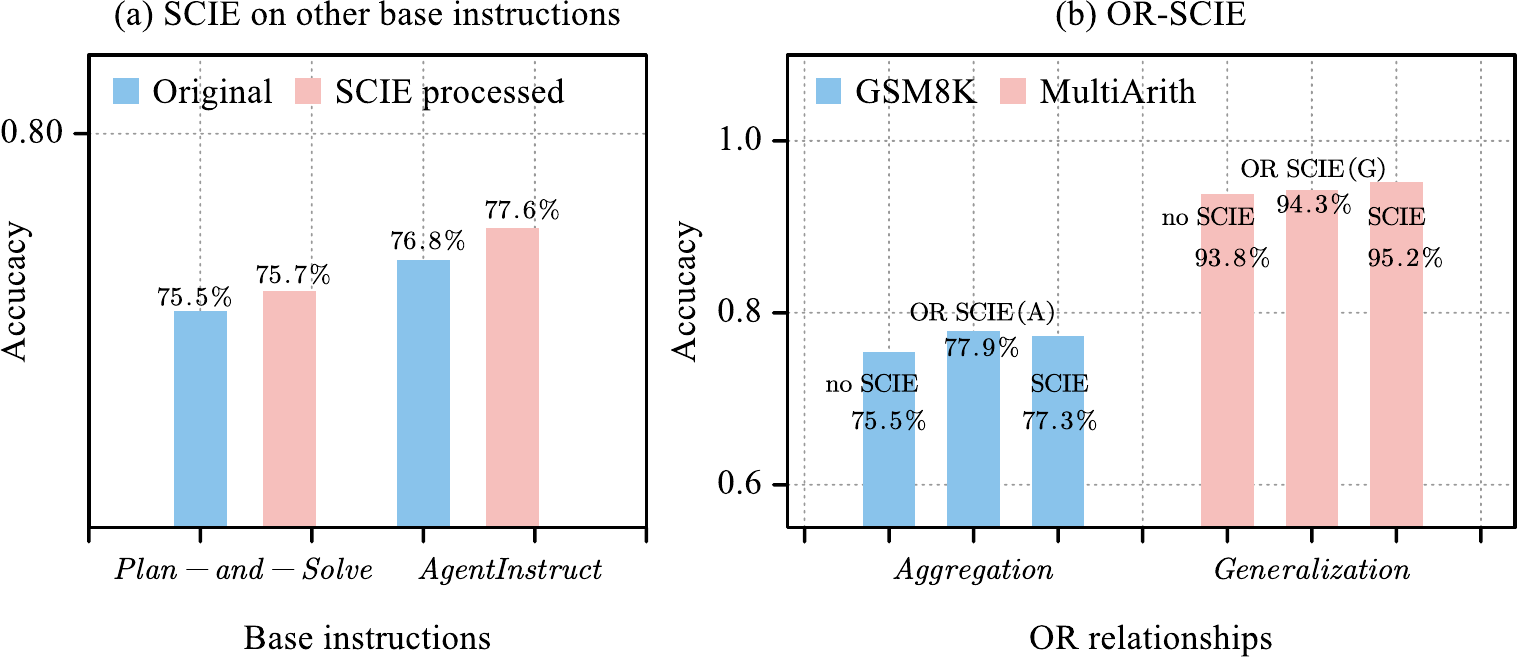}
\caption{(a) The reasoning accuracy with and without SCIE, using Plan-and-Solve and AgentInstruct as base instructions respectively. (b) The reasoning accuracy for objects inheriting causal relationships with aggregation(A) and generalization(G) OR relationships, respectively.}
\vspace{-16pt}
\label{Fig5}
\end{figure}


\subsubsection{SCIE on other base instructions.}
We evaluate the accuracy of the GSM8K of GPT-3.5 Turbo with Plan-and-Solve Prompting and AgentInstruct as the base instructions, respectively. Due to the lengthy instructions of AgentInstruct, we set $a=5$, $b=5$, for cost control. We separately conduct experiments with and without SCIE on the base instructions, obtaining results shown in Figure~\ref{Fig5} (a). The complete instructions can be found in Appendix C.

As shown in Figure~\ref{Fig5} (a), when using more complex instructions as base instructions, our method demonstrates enhancement regardless of whether the two-zero-shot prompting or one-stage prompting strategy is employed.


\subsubsection{Comparison with LLMs as Optimizers.}
LLMs as Optimizers (OPRO) give the top instructions on GPT-3.5 turbo and we will compare ours with it. Table~\ref{Tab2} shows the comparison results. Our method demonstrates better accuracy compared to OPRO. Besides, our approach requires lower training costs and offers superior interpretability.

\subsection{Evaluation of OR-SCIE}
\subsubsection{Aggregation.}
We simulate the part-whole relationship in the OR model using the GSM8K dataset as an example. Specifically, we extract the causal relationships from LLM explanations that have been proved to produce SCIE-level instructions. For instance, in this case, we extract the positive causal relationship between ``Clarity" and ``Tone" and then instruct the LLM to generate enhanced instructions based on Zero-Shot CoT (see Appendix D). Subsequently, we randomly select 70\% of the GSM8K dataset as the part object and conduct the test with instructions generated from OR-SCIE, achieving an accuracy of 77.9\% (see Figure~\ref{Fig5} (b)). Intuitively, the performance of instructions produced by the OR module may be inferior to those obtained by SCIE, as it does not optimize the ATE. In this example, the result is slightly higher than that of the complete SCIE method, which may be due to the randomly extracted test data, but it demonstrates the effectiveness of the OR module. 

\subsubsection{Generalization.}
We use the MultiArith and GSM8K datasets to simulate the generalization relationship within the OR model. MultiArith inherits the causal relationships extracted from the LLM's explanations as a class, and the resulting accuracy is 94.3\%, shown in Figure~\ref{Fig5} (b). As we can see, the results using the OR module outperform those of the no-SCIE method, and this class is reusable. To further demonstrate the practical utility of the proposed method, we evaluate GPT-4o mini and Llama-3-70b on a more challenging dataset, fresh-gaokao-math-2023 \cite{tang2024mathscale}, using the instruction directly inherited from GSM8K. Among 30 high-difficulty math problems, the instruction generated by OR-SCIE(G) enables LLMs to solve several more questions correctly compared to the base instruction.

\section{Conclusions}


This paper enhances LLMs' reasoning performance by eliciting LLMs' causal effect estimation abilities and enabling them to further self-optimize instructions. Besides, the idea of OR is introduced to achieve cost-effective reusability of the method. Experimental results not only demonstrate improved accuracy of the optimized instructions but also provide enhanced interpretability. Despite the impressive causal inference capabilities evidenced from the empirical results, exploring theoretical guarantees and refining more robust eliciting methods remain intriguing and promising avenues.

\section{Acknowledgments}
We thank the anonymous reviewers for their insightful comments. This work is supported by Beijing Normal University Zhuhai Startup Fund - Research on Artificial Intelligence Computing Models and Applications, the Beijing Normal University Zhuhai Teaching Reform Project - Online and Offline Course on Artificial Intelligence and Ethics, the Ministry of Education Supply and Demand Matching Employment-Education Integration Project: Hikvision and Beijing Normal University at Zhuhai;
Hikvision and BNU-HKBU United International College, and Guangdong Provincial Key Laboratory of Interdisciplinary Research and Application for Data Science. ZKZ and BH were supported by Guangdong Basic and Applied Basic Research Foundation Nos. 2022A1515011652 and 2024A1515012399, NSFC General Program No. 62376235, HKBU Faculty Niche Research Areas No. RC-FNRA-IG/22-23/SCI/04, and HKBU CSD Departmental
Incentive Scheme.


\bibliography{aaai25}

\begin{thebibliography}{47}
\providecommand{\natexlab}[1]{#1}

\bibitem[{Achiam et~al.(2023)Achiam, Adler, Agarwal, Ahmad, Akkaya, Aleman, Almeida, Altenschmidt, Altman, Anadkat et~al.}]{achiam2023gpt}
Achiam, J.; Adler, S.; Agarwal, S.; Ahmad, L.; Akkaya, I.; Aleman, F.~L.; Almeida, D.; Altenschmidt, J.; Altman, S.; Anadkat, S.; et~al. 2023.
\newblock Gpt-4 technical report.
\newblock \emph{arXiv preprint arXiv:2303.08774}.

\bibitem[{Bengio et~al.(2019)}]{bengio2019system}
Bengio, Y.; et~al. 2019.
\newblock From system 1 deep learning to system 2 deep learning.
\newblock In \emph{Neural Information Processing Systems}.

\bibitem[{Brown et~al.(2020)Brown, Mann, Ryder, Subbiah, Kaplan, Dhariwal, Neelakantan, Shyam, Sastry, Askell et~al.}]{brown2020language}
Brown, T.; Mann, B.; Ryder, N.; Subbiah, M.; Kaplan, J.~D.; Dhariwal, P.; Neelakantan, A.; Shyam, P.; Sastry, G.; Askell, A.; et~al. 2020.
\newblock Language models are few-shot learners.
\newblock \emph{Advances in neural information processing systems}, 33: 1877--1901.

\bibitem[{Cao et~al.(2024)Cao, Zhong, Zhou, Liu, Liu, and Han}]{cao2024envisioning}
Cao, C.; Zhong, Z.; Zhou, Z.; Liu, Y.; Liu, T.; and Han, B. 2024.
\newblock Envisioning Outlier Exposure by Large Language Models for Out-of-Distribution Detection.
\newblock In \emph{ICML}.

\bibitem[{Carey et~al.(1997)Carey, DeWitt, Naughton, Asgarian, Brown, Gehrke, and Shah}]{carey1997bucky}
Carey, M.~J.; DeWitt, D.~J.; Naughton, J.~F.; Asgarian, M.; Brown, P.; Gehrke, J.~E.; and Shah, D.~N. 1997.
\newblock The BUCKY object-relational benchmark.
\newblock In \emph{Proceedings of the 1997 ACM SIGMOD international conference on Management of data}, 135--146.

\bibitem[{Chang et~al.(2024)Chang, Xu, Wang, Luo, Xiao, and Zhu}]{chang2024efficient}
Chang, K.; Xu, S.; Wang, C.; Luo, Y.; Xiao, T.; and Zhu, J. 2024.
\newblock Efficient Prompting Methods for Large Language Models: A Survey.
\newblock \emph{arXiv preprint arXiv:2404.01077}.

\bibitem[{Cobbe et~al.(2021)Cobbe, Kosaraju, Bavarian, Chen, Jun, Kaiser, Plappert, Tworek, Hilton, Nakano et~al.}]{cobbe2021training}
Cobbe, K.; Kosaraju, V.; Bavarian, M.; Chen, M.; Jun, H.; Kaiser, L.; Plappert, M.; Tworek, J.; Hilton, J.; Nakano, R.; et~al. 2021.
\newblock Training verifiers to solve math word problems.
\newblock \emph{arXiv preprint arXiv:2110.14168}.

\bibitem[{Crispino et~al.(2024)Crispino, Montgomery, Zeng, Song, and Wang}]{crispinoagent}
Crispino, N.; Montgomery, K.; Zeng, F.; Song, D.; and Wang, C. 2024.
\newblock Agent Instructs Large Language Models to be General Zero-Shot Reasoners.
\newblock In \emph{Forty-first International Conference on Machine Learning}.

\bibitem[{Dhawan et~al.(2024)Dhawan, Cotta, Ullrich, Krishnan, and Maddison}]{dhawan2024end}
Dhawan, N.; Cotta, L.; Ullrich, K.; Krishnan, R.; and Maddison, C.~J. 2024.
\newblock End-To-End Causal Effect Estimation from Unstructured Natural Language Data.
\newblock In \emph{ICML 2024 Workshop on Foundation Models in the Wild}.

\bibitem[{Dong et~al.(2019)Dong, Yang, Wang, Wei, Liu, Wang, Gao, Zhou, and Hon}]{dong2019unified}
Dong, L.; Yang, N.; Wang, W.; Wei, F.; Liu, X.; Wang, Y.; Gao, J.; Zhou, M.; and Hon, H.-W. 2019.
\newblock Unified language model pre-training for natural language understanding and generation.
\newblock \emph{Advances in neural information processing systems}, 32.

\bibitem[{Dubey et~al.(2024)Dubey, Jauhri, Pandey, Kadian, Al-Dahle, Letman, Mathur, Schelten, Yang, Fan et~al.}]{dubey2024llama}
Dubey, A.; Jauhri, A.; Pandey, A.; Kadian, A.; Al-Dahle, A.; Letman, A.; Mathur, A.; Schelten, A.; Yang, A.; Fan, A.; et~al. 2024.
\newblock The llama 3 herd of models.
\newblock \emph{arXiv preprint arXiv:2407.21783}.

\bibitem[{Dziri et~al.(2024)Dziri, Lu, Sclar, Li, Jiang, Lin, Welleck, West, Bhagavatula, Le~Bras et~al.}]{dziri2024faith}
Dziri, N.; Lu, X.; Sclar, M.; Li, X.~L.; Jiang, L.; Lin, B.~Y.; Welleck, S.; West, P.; Bhagavatula, C.; Le~Bras, R.; et~al. 2024.
\newblock Faith and fate: Limits of transformers on compositionality.
\newblock \emph{Advances in Neural Information Processing Systems}, 36.

\bibitem[{Feder et~al.(2022)Feder, Keith, Manzoor, Pryzant, Sridhar, Wood-Doughty, Eisenstein, Grimmer, Reichart, Roberts et~al.}]{feder2022causal}
Feder, A.; Keith, K.~A.; Manzoor, E.; Pryzant, R.; Sridhar, D.; Wood-Doughty, Z.; Eisenstein, J.; Grimmer, J.; Reichart, R.; Roberts, M.~E.; et~al. 2022.
\newblock Causal inference in natural language processing: Estimation, prediction, interpretation and beyond.
\newblock \emph{Transactions of the Association for Computational Linguistics}, 10: 1138--1158.

\bibitem[{Geva et~al.(2021)Geva, Khashabi, Segal, Khot, Roth, and Berant}]{geva2021did}
Geva, M.; Khashabi, D.; Segal, E.; Khot, T.; Roth, D.; and Berant, J. 2021.
\newblock Did aristotle use a laptop? a question answering benchmark with implicit reasoning strategies.
\newblock \emph{Transactions of the Association for Computational Linguistics}, 9: 346--361.

\bibitem[{Holland(1986)}]{holland1986statistics}
Holland, P.~W. 1986.
\newblock Statistics and causal inference.
\newblock \emph{Journal of the American statistical Association}, 81(396): 945--960.

\bibitem[{Howard and Ruder(2018)}]{howard2018universal}
Howard, J.; and Ruder, S. 2018.
\newblock Universal Language Model Fine-tuning for Text Classification.
\newblock In \emph{Proceedings of the 56th Annual Meeting of the Association for Computational Linguistics (Volume 1: Long Papers)}, 328--339.

\bibitem[{Jensen(2021)}]{jensen2021improving}
Jensen, D.~D. 2021.
\newblock Improving causal inference by increasing model expressiveness.
\newblock In \emph{Proceedings of the AAAI Conference on Artificial Intelligence}, volume~35, 15053--15057.

\bibitem[{K{\i}c{\i}man et~al.(2023)K{\i}c{\i}man, Ness, Sharma, and Tan}]{kiciman2023causal}
K{\i}c{\i}man, E.; Ness, R.; Sharma, A.; and Tan, C. 2023.
\newblock Causal reasoning and large language models: Opening a new frontier for causality.
\newblock \emph{arXiv preprint arXiv:2305.00050}.

\bibitem[{Kojima et~al.(2022)Kojima, Gu, Reid, Matsuo, and Iwasawa}]{kojima2022large}
Kojima, T.; Gu, S.~S.; Reid, M.; Matsuo, Y.; and Iwasawa, Y. 2022.
\newblock Large language models are zero-shot reasoners.
\newblock \emph{Advances in neural information processing systems}, 35: 22199--22213.

\bibitem[{Kunzel et~al.(2019)Kunzel, Sekhon, Bickel, and Yu}]{kunzel2019metalearners}
Kunzel, S.~R.; Sekhon, J.~S.; Bickel, P.~J.; and Yu, B. 2019.
\newblock Metalearners for estimating heterogeneous treatment effects using machine learning.
\newblock \emph{Proceedings of the national academy of sciences}, 116(10): 4156--4165.

\bibitem[{Lee and Ogburn(2021)}]{lee2021network}
Lee, Y.; and Ogburn, E.~L. 2021.
\newblock Network dependence can lead to spurious associations and invalid inference.
\newblock \emph{Journal of the American Statistical Association}, 116(535): 1060--1074.

\bibitem[{Lester, Al-Rfou, and Constant(2021)}]{lester2021power}
Lester, B.; Al-Rfou, R.; and Constant, N. 2021.
\newblock The Power of Scale for Parameter-Efficient Prompt Tuning.
\newblock In \emph{Proceedings of the 2021 Conference on Empirical Methods in Natural Language Processing}, 3045--3059.

\bibitem[{Lewis et~al.(2019)Lewis, Liu, Goyal, Ghazvininejad, Mohamed, Levy, Stoyanov, and Zettlemoyer}]{lewis2019bart}
Lewis, M.; Liu, Y.; Goyal, N.; Ghazvininejad, M.; Mohamed, A.; Levy, O.; Stoyanov, V.; and Zettlemoyer, L. 2019.
\newblock Bart: Denoising sequence-to-sequence pre-training for natural language generation, translation, and comprehension.
\newblock \emph{arXiv preprint arXiv:1910.13461}.

\bibitem[{Li et~al.(2023)Li, Zhou, Zhu, Yao, Liu, and Han}]{li2023deepinception}
Li, X.; Zhou, Z.; Zhu, J.; Yao, J.; Liu, T.; and Han, B. 2023.
\newblock Deepinception: Hypnotize large language model to be jailbreaker.
\newblock \emph{arXiv preprint arXiv:2311.03191}.

\bibitem[{Liu et~al.(2024)Liu, Chen, Liu, Gong, Cheng, Han, and Zhang}]{liu2024discovery}
Liu, C.; Chen, Y.; Liu, T.; Gong, M.; Cheng, J.; Han, B.; and Zhang, K. 2024.
\newblock Discovery of the Hidden World with Large Language Models.
\newblock In \emph{Proceedings of the Thirty-eighth Annual Conference on Neural Information Processing Systems}.

\bibitem[{Liu et~al.(2023)Liu, Yuan, Fu, Jiang, Hayashi, and Neubig}]{liu2023pre}
Liu, P.; Yuan, W.; Fu, J.; Jiang, Z.; Hayashi, H.; and Neubig, G. 2023.
\newblock Pre-train, prompt, and predict: A systematic survey of prompting methods in natural language processing.
\newblock \emph{ACM Computing Surveys}, 55(9): 1--35.

\bibitem[{Maiya(2021)}]{maiya2021causalnlp}
Maiya, A.~S. 2021.
\newblock CausalNLP: A practical toolkit for causal inference with text.
\newblock \emph{arXiv preprint arXiv:2106.08043}.

\bibitem[{{Open Interpreter}(2024)}]{OpenInterpreter}
{Open Interpreter}. 2024.
\newblock open-interpreter: A natural language interface for computers.
\newblock https://github.com/OpenInterpreter.
\newblock Accessed: 2024-08-03.

\bibitem[{{OpenAI}(2022)}]{OpenAI2022}
{OpenAI}. 2022.
\newblock GPT-3.5 Turbo fine-tuning and API updates.
\newblock https://openai.com/index/gpt-3-5-turbo-fine-tuning-and-api-updates/.
\newblock Accessed: 2024-12-16.

\bibitem[{Prasad et~al.(2023)Prasad, Hase, Zhou, and Bansal}]{prasad2023grips}
Prasad, A.; Hase, P.; Zhou, X.; and Bansal, M. 2023.
\newblock GrIPS: Gradient-free, Edit-based Instruction Search for Prompting Large Language Models.
\newblock In \emph{Proceedings of the 17th Conference of the European Chapter of the Association for Computational Linguistics}, 3845--3864.

\bibitem[{Pryzant et~al.(2021)Pryzant, Card, Jurafsky, Veitch, and Sridhar}]{pryzant2021causal}
Pryzant, R.; Card, D.; Jurafsky, D.; Veitch, V.; and Sridhar, D. 2021.
\newblock Causal Effects of Linguistic Properties.
\newblock In \emph{Proceedings of the 2021 Conference of the North American Chapter of the Association for Computational Linguistics: Human Language Technologies}, 4095--4109.

\bibitem[{Pryzant et~al.(2023)Pryzant, Iter, Li, Lee, Zhu, and Zeng}]{pryzant2023automatic}
Pryzant, R.; Iter, D.; Li, J.; Lee, Y.~T.; Zhu, C.; and Zeng, M. 2023.
\newblock Automatic Prompt Optimization with" Gradient Descent" and Beam Search.
\newblock In \emph{The 2023 Conference on Empirical Methods in Natural Language Processing}.

\bibitem[{Pryzant et~al.(2018)Pryzant, Shen, Jurafsky, and Wagner}]{pryzant2018deconfounded}
Pryzant, R.; Shen, K.; Jurafsky, D.; and Wagner, S. 2018.
\newblock Deconfounded lexicon induction for interpretable social science.
\newblock In \emph{Proceedings of the 2018 Conference of the North American Chapter of the Association for Computational Linguistics: Human Language Technologies, Volume 1 (Long Papers)}, 1615--1625.

\bibitem[{Rosenbaum and Rubin(1983)}]{rosenbaum1983central}
Rosenbaum, P.~R.; and Rubin, D.~B. 1983.
\newblock The central role of the propensity score in observational studies for causal effects.
\newblock \emph{Biometrika}, 70(1): 41--55.

\bibitem[{Roy and Roth(2015)}]{roy2015solving}
Roy, S.; and Roth, D. 2015.
\newblock Solving General Arithmetic Word Problems.
\newblock In \emph{Proceedings of the 2015 Conference on Empirical Methods in Natural Language Processing}. Association for Computational Linguistics.

\bibitem[{Rubin(1974)}]{rubin1974estimating}
Rubin, D.~B. 1974.
\newblock Estimating causal effects of treatments in randomized and nonrandomized studies.
\newblock \emph{Journal of educational Psychology}, 66(5): 688.

\bibitem[{Suzgun et~al.(2023)Suzgun, Scales, Sch{\"a}rli, Gehrmann, Tay, Chung, Chowdhery, Le, Chi, Zhou et~al.}]{suzgun2023challenging}
Suzgun, M.; Scales, N.; Sch{\"a}rli, N.; Gehrmann, S.; Tay, Y.; Chung, H.~W.; Chowdhery, A.; Le, Q.; Chi, E.; Zhou, D.; et~al. 2023.
\newblock Challenging BIG-Bench Tasks and Whether Chain-of-Thought Can Solve Them.
\newblock In \emph{Findings of the Association for Computational Linguistics: ACL 2023}, 13003--13051.

\bibitem[{Talmor et~al.(2019)Talmor, Herzig, Lourie, and Berant}]{talmor2019commonsenseqa}
Talmor, A.; Herzig, J.; Lourie, N.; and Berant, J. 2019.
\newblock CommonsenseQA: A question answering challenge targeting commonsense knowledge.
\newblock In \emph{Proceedings of the 2019 Conference of the North}, 4149. Association for Computational Linguistics.

\bibitem[{Tang et~al.(2024)Tang, Zhang, Wang, and Wei}]{tang2024mathscale}
Tang, Z.; Zhang, X.; Wang, B.; and Wei, F. 2024.
\newblock MathScale: Scaling Instruction Tuning for Mathematical Reasoning.
\newblock In \emph{Forty-first International Conference on Machine Learning}.

\bibitem[{Wang et~al.(2023)Wang, Xu, Lan, Hu, Lan, Lee, and Lim}]{wang2023plan}
Wang, L.; Xu, W.; Lan, Y.; Hu, Z.; Lan, Y.; Lee, R. K.-W.; and Lim, E.-P. 2023.
\newblock Plan-and-Solve Prompting: Improving Zero-Shot Chain-of-Thought Reasoning by Large Language Models.
\newblock In \emph{Proceedings of the 61st Annual Meeting of the Association for Computational Linguistics (Volume 1: Long Papers)}, 2609--2634.

\bibitem[{Wang and Luo(2024)}]{wang2024exploring}
Wang, Y.; and Luo, Z. 2024.
\newblock Exploring latent discrimination through an Object-Relational Causal Inference method.
\newblock \emph{Knowledge-Based Systems}, 112148.

\bibitem[{Wei et~al.(2022)Wei, Wang, Schuurmans, Bosma, Xia, Chi, Le, Zhou et~al.}]{wei2022chain}
Wei, J.; Wang, X.; Schuurmans, D.; Bosma, M.; Xia, F.; Chi, E.; Le, Q.~V.; Zhou, D.; et~al. 2022.
\newblock Chain-of-thought prompting elicits reasoning in large language models.
\newblock \emph{Advances in neural information processing systems}, 35: 24824--24837.

\bibitem[{Wood-Doughty, Shpitser, and Dredze(2018)}]{wood2018challenges}
Wood-Doughty, Z.; Shpitser, I.; and Dredze, M. 2018.
\newblock Challenges of using text classifiers for causal inference.
\newblock In \emph{Proceedings of the Conference on Empirical Methods in Natural Language Processing. Conference on Empirical Methods in Natural Language Processing}, volume 2018, 4586. NIH Public Access.

\bibitem[{Xu et~al.(2022)Xu, Chen, Du, Shao, Yanggang, Li, and Yang}]{xu2022gps}
Xu, H.; Chen, Y.; Du, Y.; Shao, N.; Yanggang, W.; Li, H.; and Yang, Z. 2022.
\newblock GPS: Genetic Prompt Search for Efficient Few-Shot Learning.
\newblock In \emph{Proceedings of the 2022 Conference on Empirical Methods in Natural Language Processing}, 8162--8171.

\bibitem[{Yang et~al.(2023)Yang, Wang, Lu, Liu, Le, Zhou, and Chen}]{Yang2023LargeLM}
Yang, C.; Wang, X.; Lu, Y.; Liu, H.; Le, Q.~V.; Zhou, D.; and Chen, X. 2023.
\newblock Large Language Models as Optimizers.
\newblock \emph{ArXiv}, abs/2309.03409.

\bibitem[{Zhou et~al.(2022)Zhou, Muresanu, Han, Paster, Pitis, Chan, and Ba}]{zhou2022large}
Zhou, Y.; Muresanu, A.~I.; Han, Z.; Paster, K.; Pitis, S.; Chan, H.; and Ba, J. 2022.
\newblock Large Language Models are Human-Level Prompt Engineers.
\newblock In \emph{The Eleventh International Conference on Learning Representations}.

\bibitem[{Zhou et~al.(2024)Zhou, Tao, Zhu, Luo, Wang, and Han}]{zhou2024can}
Zhou, Z.; Tao, R.; Zhu, J.; Luo, Y.; Wang, Z.; and Han, B. 2024.
\newblock Can Language Models Perform Robust Reasoning in Chain-of-thought Prompting with Noisy Rationales?
\newblock In \emph{The Thirty-eighth Annual Conference on Neural Information Processing Systems}.

\end{thebibliography}

\clearpage
\appendix
\onecolumn
\section{Appendix}

\subsection{Appendix A}

The following table shows the enhanced instructions generated by LLMs on each reasoning task using Zero-Shot CoT as the base instruction. The content in the LLM explanations column is good for interpretability.

\renewcommand{\arraystretch}{1.5} 
\begin{longtable}{@{} p{2.5cm} p{10cm} p{4cm} @{}}

\toprule
\textbf{\makecell[c]{Reasoning tasks}} & \textbf{\makecell[c]{LLM explanations}} & \textbf{\makecell[c]{Enhanced instructions}} \\ 
\midrule
\endfirsthead

\toprule
\textbf{\makecell[c]{Reasoning tasks}} & \textbf{\makecell[c]{LLM explanations}} & \textbf{\makecell[c]{Enhanced instructions}} \\ 
\midrule
\endhead

\bottomrule
\endfoot

\bottomrule
\endlastfoot

GSM8K & To maximize the overall Average Treatment Effect (ATE) as evidenced by the results from the analysis, we should focus on enhancing the elements that received positive ATE values which indicate a likely benefit to the LLM's performance when these aspects are emphasized. In this case, treatments for ``Clarity" and ``Tone" have beneficial ATEs. Here's a revised instruction designed to leverage these insights: ``Ensure clarity and maintain a positive, engaging tone. Proceed methodically, articulating each thought step by step with clear, precise language." This instruction optimizes the input by emphasizing ``Clarity," which has the highest positive effect, while also considering ``Tone", fostering an engaging and precise approach meant to potentially maximize the effectiveness of the LLM’s output. & Ensure clarity and maintain a positive, engaging tone. Proceed methodically, articulating each thought step by step with clear, precise language. \\ \hline

MultiArith & To construct a new input instruction for a large language model that maximizes the overall Average Treatment Effect (ATE) based on the computed results, focus on enhancing attributes that demonstrated positive and significant influences. From our analysis, notable attributes such as 'Directness.', 'DetailLevel', and 'Urgency' demonstrated higher positive ATEs. Here is the crafted instruction integrating these key attributes: New LLM Input Instruction: "Immediately tackle the problem with a thorough and structured approach. Begin by detailing every aspect of the issue, moving sequentially from one detail to the next. Ensure each explanation is straightforward and direct, avoiding any unnecessary complexities. This methodical and focused approach will not only clarify the steps needed but also expedite effective problem-solving." This instruction integrates high-impact elements such as urgency, detail, and directness, which are theorized to maximize the effectiveness of the LLM in problem-solving tasks, reflecting our ATE findings. & Immediately tackle the problem with a thorough and structured approach. Begin by detailing every aspect of the issue, moving sequentially from one detail to the next. Ensure each explanation is straightforward and direct, avoiding any unnecessary complexities. This methodical and focused approach will not only clarify the steps needed but also expedite effective problem-solving. \\ \hline

Coin Flip & Recap of the Plan:1 Identified Treatment with Highest Positive ATE: "Directness" was found to have the highest positive ATE.2 Craft a New LLM Input Instruction: Aim to maximize overall ATE by emphasizing the 'Directness' aspect in anew LLM instruction.New LLM Instruction Generation:Given that "Directness" has the most substantial positive impact on the outcome, here's a new LLM instruction that emphasizes directness, inspired by the structure and straightforwardness of the example "Let's think step by step.":New LLM Instruction: "Directly analyze each step to promptly achieve an effective solution."This instruction is clear and straightforward, promoting a direct approach to problem-solving, which should help maximize the overall Average Treatment Effect. & Directly analyze each step to promptly achieve an effective solution. \\ \hline

Last Letters & To maximize the overall Average Treatment Effect (ATE) based on the previously generated outputs and aiming to create an instruction that synergizes well with the interaction clarity, directiveness, and focusing on the outcome while keeping the instruction straightforward, you could consider an instruction like:"Outline the solution clearly and directly, focusing on the end results. Ensure each point is concise and sequentially organized to guide comprehensively."This instruction promotes a clear, direct approach to solutions while emphasizing the importance of outcome-oriented, structured communication. It reflects higher values in clarity and directiveness, which have shown positive and significant ATE, and also encourages a focus on the outcome, which could help in enhancing the applicability and real-world relevance of the responses. & Outline the solution clearly and directly, focusing on the end results. Ensure each point is concise and sequentially organized to guide comprehensively. \\ \hline

Boolean Expressions & Given the results where "Conciseness" enhanced the Average Treatment Effect (ATE) considerably, an optimized instruction for the large language model to further increase the overall ATE would be:"Clearly and concisely step through the solution, focusing on essential details only."This instruction maintains a structured step-by-step approach while emphasizing brevity and precision, likely maximizing the positive impact on outcomes. & Concisely express the steps needed to solve the problem, focusing clearly and directly on the task. \\ \hline

StrategyQA & Given that the "Clarity" treatment produced the highest Average Treatment Effect (ATE) in our analysis,creating an instruction that emphasizes clarity is key to maximizing the overall effectiveness of the LLM in solving problems.Building on the existing example "Let's think step by step," we can enhance the instruction to further emphasize not only a methodical approach but also clarity in reasoning. This should help leverage all aspects of the model's potential by incorporating the most impactful treatment:"Let's clearly define and analyze each part of the problem step by step, ensuring clarity in each explanation and solution."This instruction encourages a detailed and transparent approach, aligning with the finding that clarity has the highest positive impact on outcomes. It should stimulate the LLM to approach problem-solving more effectively by focusing on clear, thorough explanations and step wise analysis. & Let's clearly define and analyze each part of the problem step by step, ensuring clarity in each explanation and solution. \\ \hline

CommonsenseQA & Given the analysis of the Average Treatment Effects (ATE) from the dataset, the treatments "Step-by-step guidance" and "Instructional tone" were found to have positive effects on the outcome, indicating that detailed, guided, and clearly expressed instructions potentially enhance the likelihood of eliciting correct responses in a query context like CommonsenseQA.To design a new instruction that maximizes these effects based on your example ("Let's think step by step."),and incorporating the influential attributes from the ATE results, we could use:New LLM Input Instruction: "Let's methodically analyze each possibility, step by step, ensuring our approach is clear and systematic."This instruction is crafted to:• Emphasize a sequential ("step by step") and methodical approach, leveraging the positive ATE for"Step-by-step guidance".• Use a clear and direct explanatory style, aligning with the positive ATE seen in "Instructional tone".This enhanced instruction aims to guide the LLM in engaging with the problem-solving task more efficiently and effectively. & Let's methodically analyze each possibility, step by step, ensuring our approach is clear and systematic. \\ \hline

Causal Judgement & Based on the Average Treatment Effects (ATEs) derived from the data, we should focus on constructing an instruction that enhances the highest contributing factors. Given the results: • Clarity: 0.2118 • Specificity: 0.2277 • SequentialStructure: 0.1876 These three factors have the most positive effects. The goal is to maximize these elements to achieve the highest overall ATE. Here's an example of how we might construct a new LLM input instruction to optimize the overall ATE, based on the provided example "Let's think step by step.": New LLM Input Instruction: "Clearly define each step of the process. Make sure to specifically describe the tasks involved in each step, linking them logically to ensure a smooth sequence." This instruction emphasizes: • Clarity by asking to "clearly define each step." • Specificity by ensuring detailed descriptions of "the tasks involved." • Sequential Structure by linking the steps "logically." By crafting the instruction to maximize these key factors, we enhance the likelihood of a more impactful overall ATE based on previous treatments' performances. & Clearly define each step of the process. Make sure to specifically describe the tasks involved in each step, linking them logically to ensure a smooth sequence. \\ \hline

Date Understanding & To generate an LLM input instruction that could maximize the overall Average Treatment Effect (ATE) based on the provided results, we should focus on increasing factors with positive ATE and avoid or minimize those with negative ATE. Here’s how we can tailor the instruction: 1 Include Clarity: Clarity has the highest positive ATE, so the instruction should be very clear. 2 Increase Specificity: Specific instructions should be provided, as Specificity has a positive ATE. 3 Include SequentialStructure: Using a step-by-step approach has a beneficial effect. 4 Avoid Action-Oriented terms: Since this has a negative ATE, try not to use urgent or directive language. 5 Minimize Urgency and Modularity: Both factors showed negative ATE, so these should be avoided or reduced in the instruction. Generated LLM Input Instruction: "Carefully analyze the data provided, and methodically follow each detailed step to ensure a comprehensive understanding, avoiding hurried conclusions or segmenting the analysis into unrelated parts." This instruction emphasizes clarity, specificity, and a sequential structure, while avoiding urgency and modularity, aiming to optimize the positive impact on the outcome. & Carefully analyze the data provided, and methodically follow each detailed step to ensure a comprehensive understanding, avoiding hurried conclusions or segmenting the analysis into unrelated parts. \\ \hline

DisambiguationQA & To maximize the overall Average Treatment Effect (ATE) based on the attributes we analyzed, focusing on Conciseness which had the highest positive ATE, and incorporating aspects of Urgency, the instruction could be framed to emphasize these traits while still maintaining a structured approach like the example provided. Here's a revised LLM input instruction: "Quickly outline the steps required, ensuring each is directly relevant and concisely stated." This instruction aims to maintain urgency and conciseness, which should lead to a more effective outcome based on the ATE analysis. & Quickly outline the steps required, ensuring each is directly relevant and concisely stated. \\
\end{longtable}

\clearpage
\subsection{Appendix B}
The following table shows the accuracy of reasoning tasks with and without SCIE on GPT-4o mini. We can observe that accuracy improves with SCIE for most reasoning tasks. However, an intriguing phenomenon is observed in the DisambiguationQA task, where the performance on GPT-4o mini is lower than that on GPT-3.5 Turbo. Upon examining the answers, we found that GPT-4o mini adopts a stricter definition of disambiguation, resulting in a significant number of options being labeled as 'Ambiguous'. Nevertheless, the primary focus of this paper is the impact of using SCIE on task performance rather than the models' inherent reasoning capabilities. From this perspective, SCIE remains effective on GPT-4o mini.

\begin{table}[h]
\centering
\resizebox{\textwidth}{!}{
\renewcommand{\arraystretch}{0.5}
\begin{tabular}{c c c c c c}
\midrule[1pt] 
\multirow{2}{*}{} & \multicolumn{2}{c}{Arithmetic reasoning} & \multicolumn{3}{c}{Symbolic reasoning} \\ \cmidrule(r){2-3} \cmidrule(r){4-6}
 & GSM8K & MultiArith & Coin Flip & Last Letters & Boolean Expressions \\ \midrule 
Zero-Shot CoT & 92.4 & 95.5 & 83.3 & 88.0 & 98.4 \\ \cmidrule(r){2-3} \cmidrule(r){4-6}
SCIE Zero-Shot CoT  & 92.5 & 96.3 & 84.0  & 88.8 &  97.7\\ \midrule[1pt] 
\multirow{2}{*}{} & \multicolumn{2}{c}{Common-sense reasoning} & \multicolumn{3}{c}{Other logical reasoning} \\ \cmidrule(r){2-3} \cmidrule(r){4-6}
& StrategyQA& CommonsenseQA & Causal Judgement& Date Understanding &  DisambiguationQA \\ \midrule 
Zero-Shot CoT & 76.5 & 66.9 & 63.7 & 84.5 & 48.2 \\ \cmidrule(r){2-3} \cmidrule(r){4-6}
SCIE Zero-Shot CoT  & 79.3 & 70.1 & 65.9 & 82.6 & 55.5\\ 
\midrule[1pt] 
\end{tabular}}
\end{table}

The following table shows the accuracy of reasoning tasks with and without SCIE on Llama-3-70B-T. We can observe that for most reasoning tasks, SCIE enhanced the reasoning performance.
\begin{table*}[h]
\centering
\resizebox{\textwidth}{!}{
\renewcommand{\arraystretch}{0.5}
\begin{tabular}{c c c c c c}
\midrule[1pt] 
\multirow{2}{*}{} & \multicolumn{2}{c}{Arithmetic reasoning} & \multicolumn{3}{c}{Symbolic reasoning} \\ \cmidrule(r){2-3} \cmidrule(r){4-6}
 & GSM8K & MultiArith & Coin Flip & Last Letters & Boolean Expressions \\ \midrule 
Zero-Shot CoT &  90.6 & 93.4 & 85.9 & 87.6 & 89.4 \\ \cmidrule(r){2-3} \cmidrule(r){4-6}
SCIE Zero-Shot CoT  & 91.3 & 95.2 & 86.1  & 88.2 & 88.1 \\ \midrule[1pt] 
\multirow{2}{*}{} & \multicolumn{2}{c}{Common-sense reasoning} & \multicolumn{3}{c}{Other logical reasoning} \\ \cmidrule(r){2-3} \cmidrule(r){4-6}
& StrategyQA& CommonsenseQA & Causal Judgement& Date Understanding &  DisambiguationQA \\ \midrule 
Zero-Shot CoT & 83.1 & 88.6 & 68.1 & 83.3 & 66.5 \\ \cmidrule(r){2-3} \cmidrule(r){4-6}
SCIE Zero-Shot CoT 84.5 & 85.4 & 85.9 & 68.9 & 85.7 & 69.7\\ 
\midrule[1pt] 
\end{tabular}}
\end{table*}

\clearpage
\subsection{Appendix C}
The following table shows the enhanced instructions through SCIE, using Plan-and-Sove Prompting and AgentInstruct as base instructions, separately, where the based instruction of AgentInstruct is for GSM8K specifically.
\\

\renewcommand{\arraystretch}{1.5} 
\resizebox{\textwidth}{!}{
\begin{tabular}{p{3cm} p{7cm} p{7cm}}
    \toprule
    \textbf{\makecell[c]{Prompting methods}} & \textbf{\makecell[c]{Base instructions}} & \textbf{\makecell[c]{Enhanced instructions}} \\ 
    \midrule
Plan-and-Sove Prompting & Let’s first understand the problem, extract relevant variables and their corresponding numerals, and make a plan. Then, let’s carry out the plan, calculate intermediate variables (pay attention to correct numerical calculation and commonsense), solve the problem step by step, and show the answer. & Let's decompose the problem, identify the key variables, and draft a strategy. Following that, we will execute the plan, calculate necessary intermediate variables (ensuring accuracy and logic), progressively solve the problem, and present the final solution. \\ \hline

AgentInstruct &1. Understand the question: The first step is to read the question carefully and understand what is being
asked. The questions in the GSM dataset are grade school math word problems that require multi-step
mathematical reasoning. They are designed to be solvable by a bright middle school student.
2. Identify the mathematical operations: The problems primarily involve basic arithmetic operations such as
addition, subtraction, multiplication, and division. Identify the operations required to solve the problem.
3. Apply mathematical reasoning: The problems require a sequence of elementary calculations. Apply your
mathematical reasoning to determine the sequence of steps required to solve the problem.
4. Solve the problem: Perform the calculations in the correct sequence to reach the final answer.
5. Check your work: Always check your work to ensure that your answer is reasonable and that you have not made
any calculation errors.
Tips:
- Familiarize yourself with common themes found in grade school math, such as fractions, percentages, simple
algebra, and basic geometry. These themes are likely to appear in the problems.
- Practice multi-step mathematical reasoning. The problems in the GSM dataset require between 2 and 8 steps to
solve.
- Remember that the problems are designed to be solvable by a bright middle school student. If a problem
seems overly complex, you may be overthinking it. Try to simplify the problem and break it down into smaller,
manageable steps.& 1. Structure of the Dataset: The GSM8K dataset comprises grade school math word problems. Each instance consists of a string for the math question and a string for the corresponding answer, with step-by-step reasoning and calculator annotations.

2. Understanding Math Problems: Problems necessitate multi-step reasoning involving basic arithmetic operations like addition, subtraction, multiplication, and division.

3. Answer Format: Each problem's answer includes the reasoning steps and final answer. Steps are annotated with calculations.

4. Solving the Questions:
   - Grasp the problem requirements.
   - Outline the necessary steps for the solution.
   - Execute step-by-step calculations with annotations.
   - Present the final answer.

5. Mathematical Topics: Familiarity with basic arithmetic operations and multi-step reasoning is essential.

6. Problem-solving Steps:
   - Carefully read and comprehend the question.
   - Identify the required steps for the solution.
   - Annotate calculations for each step.
   - State the final answer.

Remember, the dataset targets problems solvable by a bright middle school student using basic arithmetic knowledge. Abide by these guidelines to tackle each problem in the GSM8K dataset effectively. \\ 
    \bottomrule
\end{tabular}}

\vspace{15pt}
\subsection{Appendix D}
The following table shows the enhanced instruction generated from inheriting the causal relationship as class based on "Let's think step by step.".
\\

\renewcommand{\arraystretch}{1.5} 
\begin{tabular}{@{} p{2cm} p{4cm} p{11cm} @{}}
    \toprule
    \textbf{\makecell[c]{OR module}} & \textbf{\makecell[c]{Base instruction}} & \textbf{\makecell[c]{Enhanced instruction}} \\ 
    \midrule
\makecell[c]{OR-SCIE} & \makecell[c]{Let’s think step by step.} & Please provide a clear, step-by-step solution to the following problem. Go directly to the steps and final answer without any introductory or concluding remarks. \\ 
    \bottomrule
\end{tabular}









\clearpage
\subsection{Appendix E}
We conduct an ablation study about the counterfactual instruction generation in the data generation part of SCIE. We directly generate the same number $n(a+1)$ of instructions without counterfactual instructions according to each proxy feature, and keep the other steps unchanged. The experimental setting is Zero-Shot CoT as the base instruction, $a=9$, $b=5$, and tested on GPT-3.5 Turbo. The following figures show the ATE estimation of the SCIE without the counterfactual module (left), and the accuracy of different methods (right). 

\begin{figure}[h]
\centering
\includegraphics[width=0.7\textwidth]{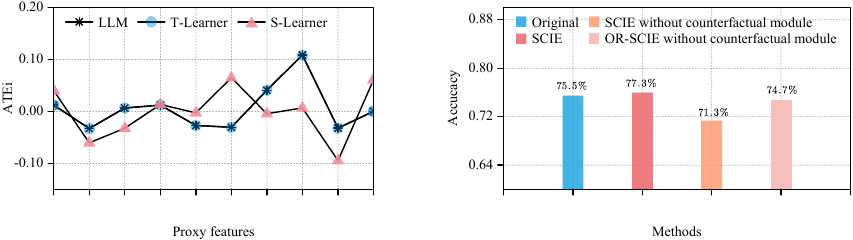}
\label{AppE}
\end{figure}

We can observe that differences in the ATE estimation between the LLM (T-Learner) and S-Learner, indicating that the causal effects estimated by the LLM are not be stable and robust. This is further evidenced by the right graph, which shows that the accuracy achieved with SCIE without the counterfactual module is lower than that obtained without enhancement methods. This demonstrates that the absence of counterfactual steps in observational data significantly impacts data quality, adversely affecting subsequent causal effect estimation and instruction enhancement. The performance of OR-SCIE without the counterfactual module is better than SCIE without it, suggesting that the OR module requires enhanced ATE estimation to ensure performance stability.

The following table shows the instructions with LLM explanations of the (OR)-SCIE without the counterfactual module.
\\

\renewcommand{\arraystretch}{1.5} 
\begin{tabular}{@{} p{2cm} p{8cm} p{7cm} @{}}
    \toprule
    \textbf{\makecell[c]{Methods}} & \textbf{\makecell[c]{LLM explanations}} & \textbf{\makecell[c]{Instructions}} \\ 
    \midrule
SCIE without counterfactual module & To optimize the overall ATE, the instruction should encourage behaviors or characteristics associated with the treatments that had positive effects based on the calculated ATE results. Positive effects were noted in the following treatments from your causal analysis:• Clarity• Specificity• Sequentiality• Directness• Physical Breakdown of Tasks• Encouragement of Immediate Action Given that "Physical Breakdown of Tasks" showed the highest ATE, more emphasis should be placed on this aspect.Here's a revised version of your example instruction that incorporates these successful elements, specifically designed to maximize the overall ATE:"Let’s clearly outline every task required, step by step, ensuring each directive is specific and directly linked to the immediate goal. Focus on breaking down the tasks physically where needed, and encourage prompt action to facilitate effective completion."This instruction leverages aspects that have proven to have a positive impact on outcomes, making the large language model's instruction more effective for achieving a higher Average Treatment Effect. & Let’s clearly outline every task required, step by step, ensuring each directive is specific and directly linked to the immediate goal. Focus on breaking down the tasks physically where needed, and encourage prompt action to facilitate effective completion. \\ \hline

OR-SCIE without counterfactual module & Instruction: "Break down the problem into clear, manageable steps. Start with the first step and provide a direct answer without any additional commentary. Move sequentially through each step until the solution is fully articulated. Focus on clarity and specificity in each response, ensuring that each task is outlined explicitly. Encourage immediate action at each stage. Let's begin!" & Break down the problem into clear, manageable steps. Start with the first step and provide a direct answer without any additional commentary. Move sequentially through each step until the solution is fully articulated. Focus on clarity and specificity in each response, ensuring that each task is outlined explicitly. Encourage immediate action at each stage. Let's begin! \\ 
    \bottomrule
\end{tabular}



\end{document}